\documentclass[11pt,a4paper]{article}
\usepackage[hyperref]{acl2021}
\usepackage{times}
\usepackage{latexsym}
\usepackage{subfigure}

\usepackage{multirow}
\usepackage{graphicx}
\usepackage{microtype}
\usepackage{booktabs}
\usepackage{amsmath}
\usepackage{multirow}
\usepackage{color}
\usepackage{footmisc}
\usepackage{soul}
\usepackage{arydshln}
\usepackage{graphics}
\usepackage{graphicx}
\usepackage{algorithm}
\usepackage{xurl}
\usepackage{algpseudocode}
\definecolor{lightgreen}{HTML}{CEF6CE}
\definecolor{lightred}{HTML}{FD8F70}

\aclfinalcopy 
\def\aclpaperid{1023} 

\title{Modeling Discriminative Representations for Out-of-Domain Detection with Supervised Contrastive Learning}

\author{Zhiyuan Zeng$^{1*}$, Keqing He$^{2*}$, Yuanmeng Yan$^{1}$ , Zijun Liu$^{1}$, {\bf Yanan Wu$^{1}$} \\ {\bf Hong Xu$^{1}$,} {\bf Huixing Jiang$^{2}$,} {\bf Weiran Xu$^{1}$}\thanks{\ \ The first two authors contribute equally. Weiran Xu is the corresponding author.}\\
 $^1$Pattern Recognition \& Intelligent System Laboratory \\
  $^1$Beijing University of Posts and Telecommunications, Beijing, China\\
$^{2}$Meituan Group, Beijing, China\\
  \texttt{\{zengzhiyuan,yanyuanmeng,liuzijun,yanan.wu,xuhong\}@bupt.edu.cn}\\
  \texttt{\{hekeqing,jianghuixing\}@meituan.com},
  \texttt{\{xuweiran\}@bupt.edu.cn}
  }
\date{}
\begin{document}
\maketitle
\begin{abstract}
Detecting Out-of-Domain (OOD) or unknown intents from user queries is essential in a task-oriented dialog system. A key challenge of OOD detection is to learn discriminative semantic features. Traditional cross-entropy loss only focuses on whether a sample is correctly classified, and does not explicitly distinguish the margins between categories. In this paper, we propose a supervised contrastive learning objective to minimize intra-class variance by pulling together in-domain intents belonging to the same class and maximize inter-class variance by pushing apart samples from different classes. Besides, we employ an adversarial augmentation mechanism to obtain pseudo diverse views of a sample in the latent space. Experiments on two public datasets prove the effectiveness of our method capturing discriminative representations for OOD detection. \footnote{Our code is available at \url{https://github.com/parZival27/supervised-contrastive-learning-for-out-of-domain-detection}.}
\end{abstract}

\section{Introduction}





Detecting Out-of-Domain (OOD) or unknown intents from user queries is an essential component in a task-oriented dialog system \cite{Gnewuch2017TowardsDC,Akasaki2017ChatDI,Tulshan2018SurveyOV,Shum2018FromET}. It aims to know when a user query falls outside their range of predefined supported intents to avoid performing wrong operations. 
Different from normal intent detection tasks, we do not know the exact number of unknown intents in practical scenarios and can barely annotate extensive OOD samples. Lack of real OOD examples leads to poor prior knowledge about these unknown intents, making it challenging to identify OOD samples in the task-oriented dialog system.

Previous methods of OOD detection can be generally classified into two types: supervised and unsupervised OOD detection. Supervised OOD detection \cite{Scheirer2013TowardOS,Fei2016BreakingTC,Kim2018JointLO,larson-etal-2019-evaluation,Zheng2020OutofDomainDF,9413908} represents that there are extensive labeled OOD samples in the training data. In contrast, unsupervised OOD detection \cite{Bendale2016TowardsOS,Hendrycks2017ABF,Shu2017DOCDO,Lee2018ASU,Ren2019LikelihoodRF,Lin2019DeepUI,xu-etal-2020-deep,Zeng2021AdversarialSL} means no labeled OOD samples except for labeled in-domain data. Specifically, for supervised OOD detection, \newcite{Fei2016BreakingTC,larson-etal-2019-evaluation}, form a $(N+1)$-class classification problem where the $(N+1)$-th class represents the unseen intents. 
Further, \newcite{Zheng2020OutofDomainDF} uses labeled OOD data to generate an entropy regularization term to enforce the predicted distribution of OOD inputs closer to the uniform distribution. 
However, these methods heavily rely on large-scale time-consuming labeled OOD data. Compared to these supervised methods, unsupervised OOD detection first learns discriminative intent representations via in-domain (IND) data, then employs detecting algorithms, such as Maximum Softmax Probability (MSP) \cite{Hendrycks2017ABF},  Local Outlier Factor (LOF) \cite{Lin2019DeepUI},  Gaussian Discriminant Analysis (GDA) \cite{xu-etal-2020-deep} to compute the similarity of features between OOD samples and IND samples. 
In this paper, we focus on the unsupervised OOD detection.

\begin{figure*}[t]
    \centering
    \setlength{\abovecaptionskip}{-0.01cm}
    \resizebox{0.75\textwidth}{!}{
    \includegraphics{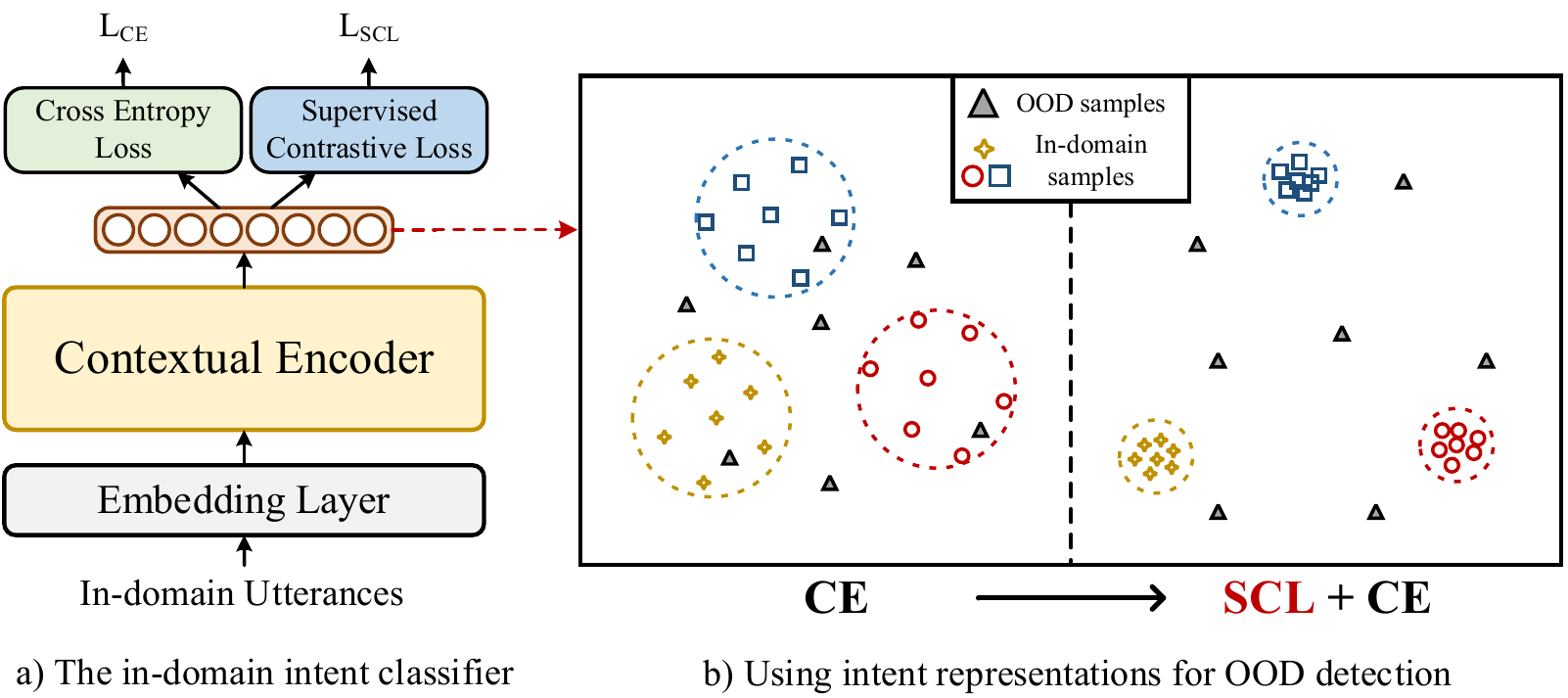}}
    \caption{The overall architecture of our proposed method. We first train an intent classifier on IND data using CE or SCL+CE objectives. Then, we extract the intent representation of a test sample to detect OOD.}
    \label{fig:model}
    \vspace{-0.6cm}
\end{figure*}

A key challenge of unsupervised OOD detection is to learn discriminative semantic features via IND data. We hope to cluster the same type of IND intents more tightly and separate different types of IND intents further. Traditional softmax loss \cite{Hendrycks2017ABF} only focuses on whether the sample is correctly classified, and does not explicitly distinguish the relationship between categories. Further, \newcite{Lin2019DeepUI} proposes a large margin cosine loss (LMCL) \cite{Wang2018AdditiveMS} which
maximizes the decision margin in the latent space. LMCL forces the model to not only classify correctly but also maximize inter-class variance and minimize intra-class variance. Following the similar motivation, we aim to pull intents belonging to the same class together while simultaneously pushing apart samples from different classes to further model discriminative semantic features.

In this paper, we propose a supervised contrastive learning (SCL) model to learn discriminative semantic intent representation for OOD detection. SCL aims to minimize intra-class variance by pulling together IND intents belonging to the same class and maximize inter-class variance by pushing apart samples from different classes. Empirical results demonstrate the effectiveness of discriminative representation for OOD detection. Besides, to enhance the diversity of data augmentation in SCL, we employ an adversarial attack mechanism to obtain pseudo hard positive samples in the latent space by computing model-agnostic adversarial worst-case perturbations to the inputs. Our contributions are three-fold: (1) To the best of our knowledge, we are the first to apply supervised contrastive learning to OOD detection. (2) Compared to cross-entropy (CE) loss, SCL+CE can maximize inter-class variance and minimize intra-class variance to learn discriminative semantic representation. (3) Extensive experiments and analysis on two public datasets demonstrate the effectiveness of our method.

\begin{table*}[t]
\centering
\resizebox{0.80\textwidth}{!}{%
\begin{tabular}{l|l|cc|cc|cc|cc}
\hline
\multicolumn{2}{c|}{\multirow{3}{*}{Models}} & \multicolumn{4}{c|}{CLINC-Full} & \multicolumn{4}{c}{CLINC-Small} \\ \cline{3-10} 
\multicolumn{2}{c|}{} & \multicolumn{2}{c|}{IND} & \multicolumn{2}{c|}{OOD} & \multicolumn{2}{c|}{IND} & \multicolumn{2}{c}{OOD} \\ \cline{3-10} 
\multicolumn{2}{c|}{} & ACC & F1 & Recall & F1 & ACC & F1 & Recall & F1 \\ \hline
\multirow{4}{*}{LSTM} & CE & 86.34 & 87.73 & 63.72 & 65.23 & 84.24 & 84.30 & 60.40 & 61.07 \\ 
 & LMCL & 86.83 & 87.90 & 64.14 & 65.79 & 84.46 & 84.87 & 60.72 & 61.89 \\ \cline{2-10} 
 & SCL+CE(ours) & 87.01 & 88.28 & 66.80 & 67.68 & 85.73 & 86.61 & 63.96 & 64.44 \\ 
 & SCL+LMCL(ours) & \textbf{87.37} & \textbf{88.60} & \textbf{66.92} & \textbf{68.04} & \textbf{85.93} & \textbf{87.02} & \textbf{64.16} & \textbf{64.70} \\ \hline\hline
\multirow{4}{*}{BERT} & CE & 88.13 & 88.98 & 64.24 & 66.17 & 86.68 & 86.20 & 61.64 & 62.58 \\ 
 & LMCL & 88.57 & 89.12 & 64.76 & 66.80 & 86.76 & 86.64 & 62.20 & 63.11 \\ \cline{2-10} 
 & SCL+CE(ours) & 88.97 & 89.57 & 66.84 & 68.03 & 87.65 & 88.07 & 64.44 & 64.52 \\ 
 & SCL+LMCL(ours) & \textbf{89.20} & \textbf{90.03} & \textbf{67.28} & \textbf{68.21} & \textbf{87.87} & \textbf{88.30} & \textbf{64.64} & \textbf{65.01} \\ \hline
\end{tabular}%
}
\vspace{-0.2cm}
\caption{Performance comparison on CLINC-Full and CLINC-Small datasets ($p < 0.05$ under t-test).}
\vspace{-0.6cm}
\label{tab1}
\end{table*}

\section{Methodology}

\textbf{Overall Architecture} Fig \ref{fig:model} shows the overall architecture of our proposed method. As Fig \ref{fig:model}(a) displays, we first train an IND intent classifier using CE or SCL+CE objectives in the training stage. Then in the test stage, we extract the intent feature of a test sample and employ the detection algorithms MSP \cite{Hendrycks2017ABF}, LOF \cite{Lin2019DeepUI} or GDA \cite{xu-etal-2020-deep} to detect OOD. \footnote{In this paper, we focus on the first training stage. Thus we dive into the details about the detection algorithms MSP, LOF and GDA in the appendix.} Fig \ref{fig:model}(b) demonstrates the effectiveness of our method capturing discriminative intent representations, where SCL+CE can maximize inter-class variance and minimize intra-class variance.

\textbf{Supervised Contrastive Learning} 
We first review the classic cross-entropy (CE) loss and its improved version, large margin cosine loss (LMCL). Then we explain our supervised contrastive loss (SCL) in detail. Given an IND sample $x_{i}$ and its intent label $y_{i}$, we adopt a BiLSTM \cite{Hochreiter1997LongSM} or BERT \cite{Devlin2019BERTPO} encoder to get the intent representation $s_{i}$. The CE loss and LMCL are defined as follows \footnote{For brevity, we omit the L2 normalization on both features and weight vectors for LMCL.}:
{\setlength{\abovedisplayskip}{0.1cm}
\setlength{\belowdisplayskip}{0.1cm}
\begin{align}
&\mathcal{L}_{CE} = 
\frac{1}{N} \sum_{i}-\log \frac{e^{W_{y_{i}}^{T} s_{i}/\tau}}{\sum_{j} e^{W_{j}^{T} s_{i}/\tau}} \\
&\mathcal{L}_{LMCL}\!=\!\frac{1}{N}\!\sum_{i}\!-\log\!\frac{e^{W_{y_{i}}^{T} s_{i}/\tau}}{e^{W_{y_{i}}^{T} s_{i}/\tau}\!+\!\sum_{j \neq y_{i}} e^{(W_{j}^{T} s_{i}+m)/\tau}} 
\end{align}}
where $N$ denotes the number of training samples, $y_{i}$ is the ground-truth class of the $i$-th sample, $\tau$ is the temperature factor, $W_{j}$ is the weight vector of the $j$-th class, and $m$ is the cosine margin. Compared to CE, LMCL adds a normalized decision margin on the negative classes and forces the model to explicitly distinguish positive class and negative classes. Our experiment \ref{main_result} shows LMCL can slightly improve the performance of OOD detection. To further model discriminative intent representations, motivated by recent contrastive learning work \cite{Chen2020ASF,He2020MomentumCF,Khosla2020SupervisedCL,Gunel2020SupervisedCL}, we propose a supervised contrastive learning objective to minimize intra-class variance and  maximize inter-class variance:
{\setlength{\abovedisplayskip}{0.05cm}
\setlength{\belowdisplayskip}{0.1cm}
\begin{align}
\begin{split}
    \mathcal{L}_{S C L}=&\sum_{i=1}^{N}-\frac{1}{N_{y_{i}}-1} \sum_{j=1}^{N} \mathbf{1}_{i \neq j} \mathbf{1}_{y_{i}=y_{j}} \\ &\log \frac{\exp \left(s_{i} \cdot s_{j} / \tau\right)}{\sum_{k=1}^{N} \mathbf{1}_{i \neq k} \exp \left(s_{i}\cdot s_{k} / \tau\right)}
\end{split}
\end{align}}
where $N_{y_{i}}$ is the total number of examples in the batch that have the same label as $y_{i}$ and $\mathbf{1}$ is an indicator function. Note that we only perform SCL on the IND data since we focus on the unsupervised OOD detection where no labeled OOD data exists. As Fig \ref{fig:model}(b) shows, SCL aims to pull together IND intents belonging to the same class and pushing apart samples from different classes, which helps recognize OOD intents near the decision boundary. In the implementation, we first pre-train the intent classifier using SCL, then finetune the model using CE or LMCL, both on the IND data. We compare iterative training and joint training in the appendix.

\textbf{Adversarial Augmentation} \newcite{Chen2020ASF} has proved the necessity of data augmentation for contrastive learning. However, there is no simple and effective augmentation strategy in the NLP area, which requires much handcrafted engineering. Thus, we apply adversarial attack \cite{goodfellow2014explaining, kurakin2016adversarial, jia2017adversarial, zhang2019generating,Yan2020AdversarialSD} to generate pseudo positive samples to increase the diversity of views for contrastive learning. Specifically, we need to compute the worst-case perturbation $\boldsymbol{\delta}$ that maximizes the original cross-entropy loss $\mathcal{L}_{CE}$: $\boldsymbol{\delta}=\underset{\left\|\boldsymbol{\delta}^{\prime}\right\| \leq \epsilon}{\arg \max } \quad \mathcal{L}_{CE} \left(\boldsymbol{\theta}, \boldsymbol{x}+\boldsymbol{\delta}^{\prime}\right)$, where $\boldsymbol{\theta}$ represents the parameters of the intent classifier and $\boldsymbol{x}$ denotes a given sample. $\epsilon$ is the norm bound of the perturbation $\boldsymbol{\delta}$. We apply Fast Gradient Value (FGV) \cite{rozsa2016adversarial} to approximate the perturbation $\boldsymbol{\delta}$: 
{\setlength{\abovedisplayskip}{0.05cm}
\setlength{\belowdisplayskip}{0.1cm}
\begin{align}
    \boldsymbol{\delta} = \epsilon \frac{g}{||g||}; \text{where } g=\nabla_{\boldsymbol{x}} \mathcal{L}_{CE} (f(\boldsymbol{x};\boldsymbol{\theta}),y)
\end{align}}
We perform normalization to $g$ and then use a small $\epsilon$ to ensure the approximate is reasonable. Finally, we can obtain the pseudo augmented sample $\boldsymbol{x}_{adv} = \boldsymbol{x} + \boldsymbol{\delta}$ in the latent space. The pseudo samples are applied to augment positive views per anchor in SCL. Ablation study \ref{noise} shows adversarial augmentation significantly improves the performance of SCL for OOD detection.

\section{Experiments}
\subsection{Setup}

\textbf{Datasets} We use two benchmark OOD datasets, CLINC-Full and CLINC-Small \cite{larson-etal-2019-evaluation}. We report IND metrics: Accuracy(Acc) and F1, and OOD metrics: Recall and F1. OOD Recall and F1 are the main evaluation metrics in this paper. \textbf{Baselines} We adopt LSTM and BERT as our intent classifier and compare SCL with CE and LMCL. Since only using SCL can't classify in-domain intents directly, we first pre-train the classifier using SCL, then finetune the model using CE or LMCL, both on the IND data. We use three OOD detection algorithms MSP, LOF and GDA to verify the generalization capability of SCL. We present dataset statistics, implementation details, and results on MSP and LOF in the appendix.

\subsection{Main Results}
\label{main_result}

Tab \ref{tab1} displays the main results on GDA. Combining SCL and CE/LMCL significantly outperforms all the baselines, both on OOD and IND metrics. For OOD metrics, using SCL+CE in LSTM outperforms CE by 3.08\%(Recall) and 2.45\%(F1) on CLINC-Full, 3.56\%(Recall) and 3.37\%(F1) on CLINC-Small. Similar improvements based on LMCL are observed. The results prove the effectiveness of SCL for OOD detection. For IND metrics, using SCL+CE in LSTM outperforms CE by 0.67\%(ACC) and 0.55\%(F1) on CLINC-Full, 1.49\%(ACC) and 2.31\%(F1) on CLINC-Small. The results confirm SCL also helps IND intent detection. The difference between OOD and IND improvements is probably attributed to metric scale and data imbalance in the original test set. Besides, SCL gains higher improvements on CLINC-Small than CLINC-Full, which displays the advantage of our approach in the few-shot scenario (see details in Section \ref{few-shot}). SCL also gets consistent improvements on BERT by 2.60\%(Recall) and 1.86\%(F1) on CLINC-Full OOD metrics, 0.84\%(ACC) and 0.59\%(F1) on CLINC-Full IND metrics,
substantiating our method is model-agnostic for different OOD detection architectures.

\begin{table}[t]
\centering
\resizebox{0.48\textwidth}{!}{%
\begin{tabular}{l|l|c|c|c|c}
\hline
\multicolumn{2}{c|}{models} & min & max & mean & median \\ \hline
\multirow{2}{*}{LSTM} & CE & 1.13E-07 & 2.63E-04 & 4.23E-05 & 1.61E-05 \\ \cline{2-6} 
 & SCL+CE & 4.35E-08 & 1.85E-04 & 3.23E-05 & 1.39E-05 \\ \hline\hline
\multirow{2}{*}{BERT} & CE & 8.26E-08 & 2.23E-04 & 3.84E-05 & 1.56E-05 \\ \cline{2-6} 
 & SCL+CE & 2.86E-08 & 1.67E-04 & 3.05E-05 & 1.36E-05 \\ \hline
\end{tabular}%
}
\vspace{-0.2cm}
\caption{Intra-class variance statistics.}
\vspace{-0.5cm}
\label{tab2}
\end{table}

\begin{figure}[t]
    \centering
    \resizebox{.52\textwidth}{!}{
    \includegraphics{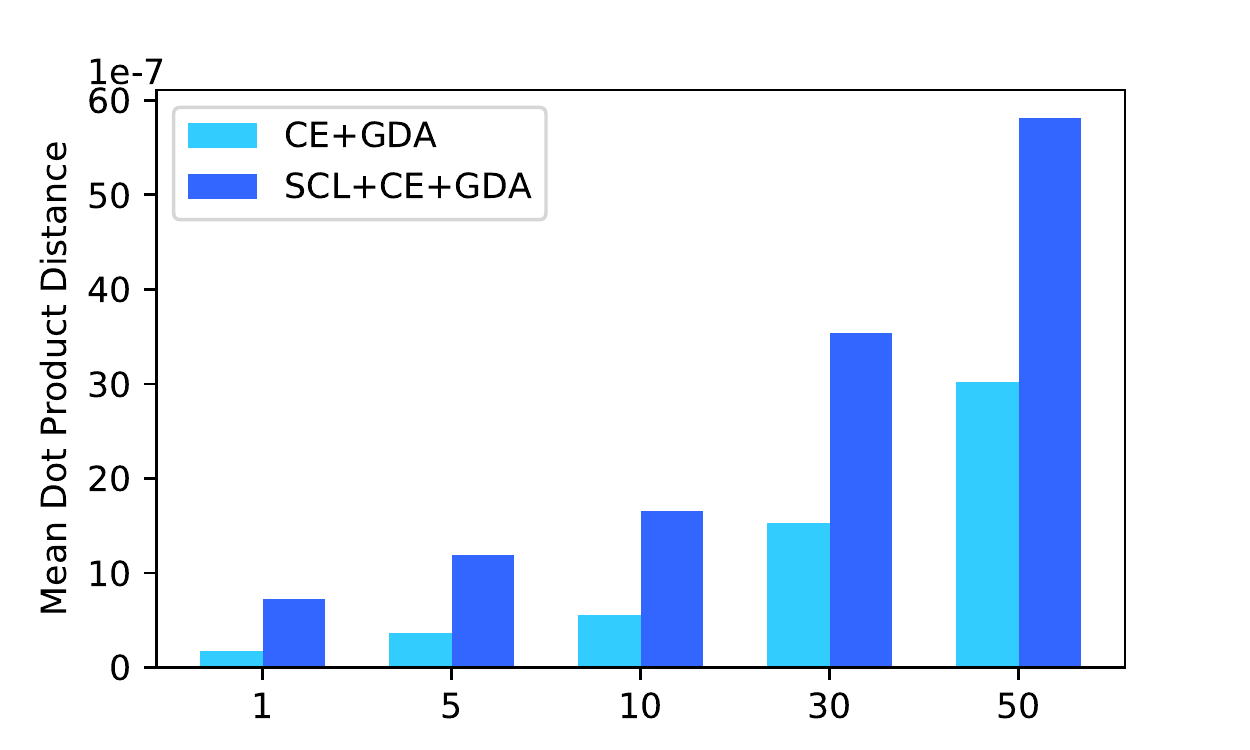}}
    \vspace{-0.7cm}
    \caption{Comparison between inter-class distances.}
    \label{fig2}
     \vspace{-0.7cm}
\end{figure}

\subsection{Analysis}

\noindent\textbf{Analysis of IND feature distribution.} We analyze the representation distribution of IND data on CLINC-Full dataset from two perspectives, intra-class and inter-class. We choose SCL+CE based on GDA to perform analysis. Tab \ref{tab2} shows the statistics of intra-class variance, which can indicate the degree of clustering of intra-class data representations. Specifically, we average the variances of each sample normalized representation with the same intent label to its cluster center in the test set as cluster intra-class variance, then report min/max/mean/median values on all cluster intra-class variances. Results show SCL effectively decreases intra-class variances, especially in terms of max and mean values, which confirms SCL can converge intra-class intent representations. 

Fig \ref{fig2} shows the inter-class distances. We average dot product distances between each class center to its k nearest class centers, then average results of all classes as inter-class distance. The X-axis denotes the number of k. We observe a significant increase in SCL+CE compared to CE. When k is smaller, the increase is more obvious. It verifies SCL can maximize inter-class variance and distinguish intent classes. We also provide visualization analysis in the appendix. In summary, SCL can pull together IND intents belonging to the same class and push apart samples from different classes, which makes representations more discriminative.

\begin{table}[t]
\resizebox{0.48\textwidth}{!}{%
\begin{tabular}{ll|lllll}
\hline
\multicolumn{2}{l|}{Proportion} & 10\% & 20\% & 30\% & 40\% & 50\% \\ \hline
\multicolumn{1}{l|}{\multirow{3}{*}{IND F1}} & CE & 63.31 & 70.77 & 77.84 & 81.55 & 84.30 \\ \cline{2-7} 
\multicolumn{1}{l|}{} & SCL+CE & 69.50 & 75.14 & 81.45 & 84.18 & 86.61 \\ \cline{2-7} 
\multicolumn{1}{l|}{} & $\mathit{Relative} \mathbf{\uparrow}$ & \textbf{9.78\%} & \textbf{6.17\%} & \textbf{4.64\%} & \textbf{3.23\%} & \textbf{2.74\%} \\ \hline
\multicolumn{1}{l|}{\multirow{3}{*}{OOD F1}} & CE & 42.16 & 48.34 & 53.00 & 57.92 & 61.07 \\ \cline{2-7} 
\multicolumn{1}{l|}{} & SCL+CE & 50.10 & 54.43 & 58.61 & 62.12 & 64.44 \\ \cline{2-7} 
\multicolumn{1}{l|}{} & $\mathit{Relative} \mathbf{\uparrow}$ & \textbf{18.83\%} & \textbf{12.60\%} & \textbf{10.58\%} & \textbf{7.25\%} & \textbf{5.52\%} \\ \hline
\end{tabular}%
}
\vspace{-0.2cm}
\caption{Effect of training data size.}
\vspace{-0.3cm}
\label{tab3}
\end{table}

\begin{figure}[t]
    \centering
    \resizebox{.50\textwidth}{!}{
    \includegraphics{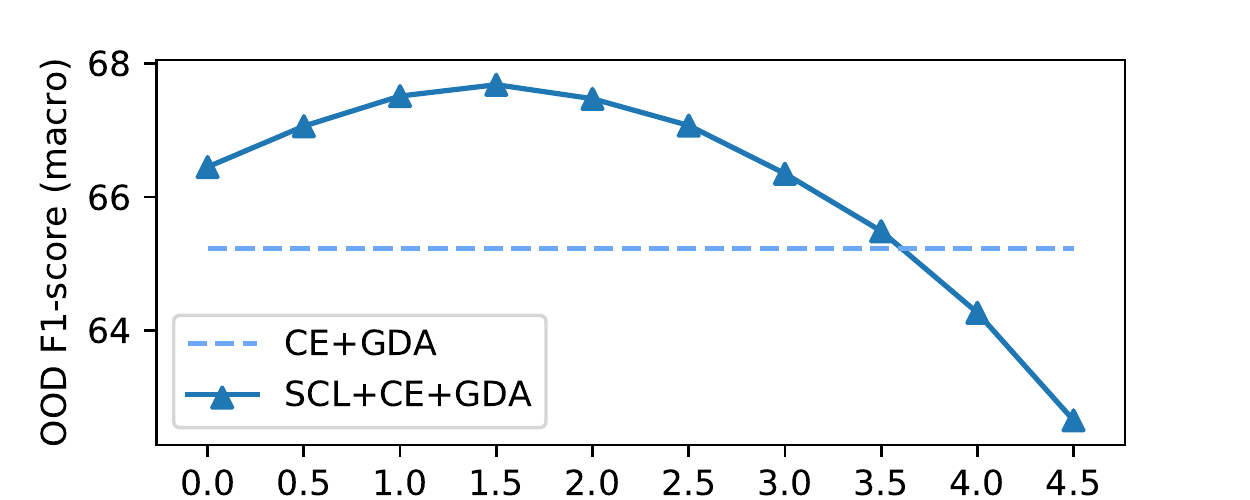}}
    \vspace{-0.6cm}
    \caption{Effect of Adversarial Perturbation Norm.}
    \label{fig3}
     \vspace{-0.6cm}
\end{figure}

\begin{figure*}[t]
\begin{minipage}[t]{0.33\linewidth}
\centering
\includegraphics[width=2.2in]{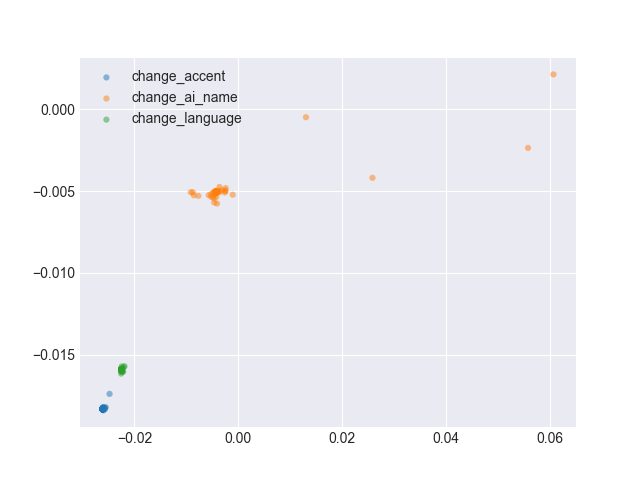}
\label{figa1}
\end{minipage}%
\begin{minipage}[t]{0.33\linewidth}
\centering
\includegraphics[width=2.2in]{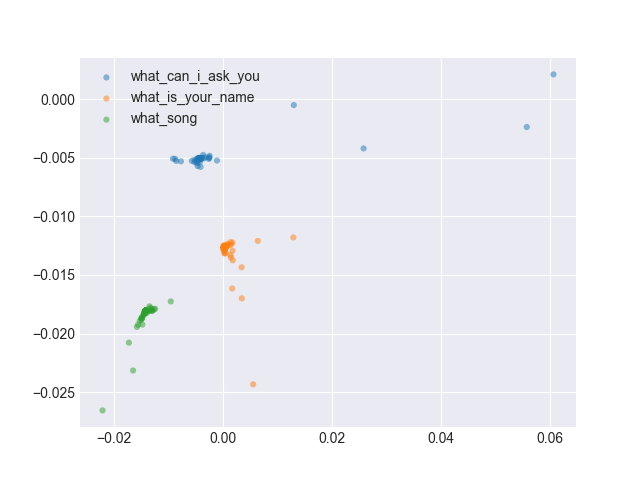}
\label{figb1}
\end{minipage}
\begin{minipage}[t]{0.33\linewidth}
\centering
\includegraphics[width=2.2in]{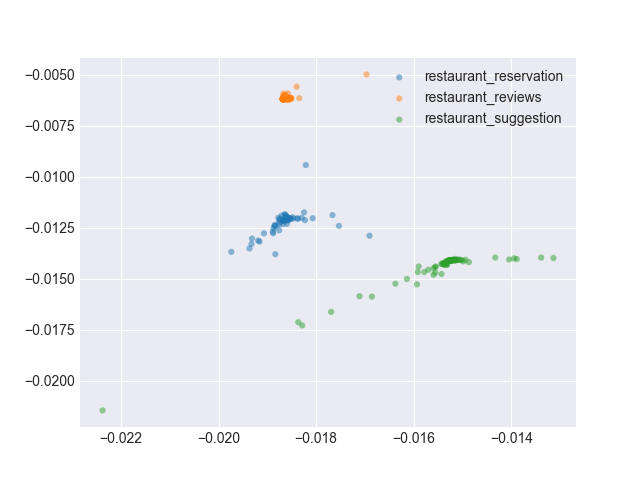}
\label{figc1}
\end{minipage}

\vspace{-0.8cm}
\begin{minipage}[t]{0.33\linewidth}
\centering
\includegraphics[width=2.2in]{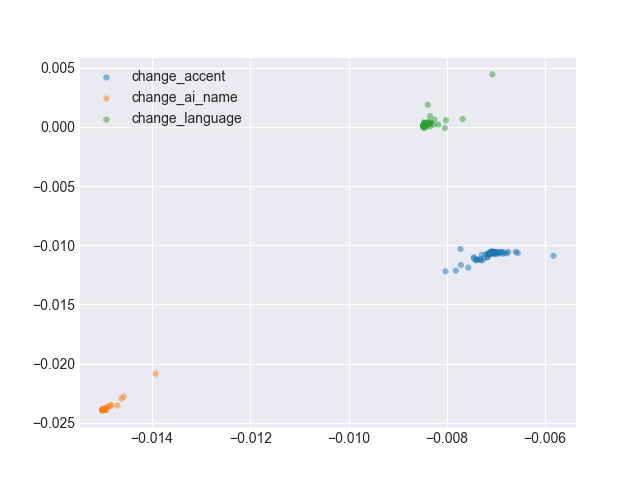}
\label{figa2}
\end{minipage}%
\begin{minipage}[t]{0.33\linewidth}
\centering
\includegraphics[width=2.2in]{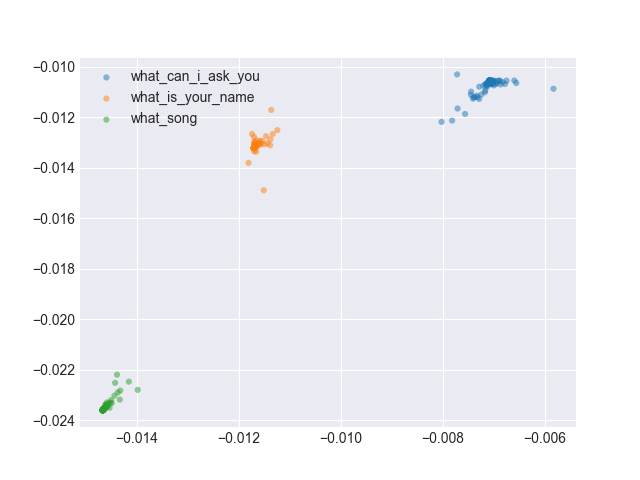}
\label{figb2}
\end{minipage}
\begin{minipage}[t]{0.33\linewidth}
\centering
\includegraphics[width=2.2in]{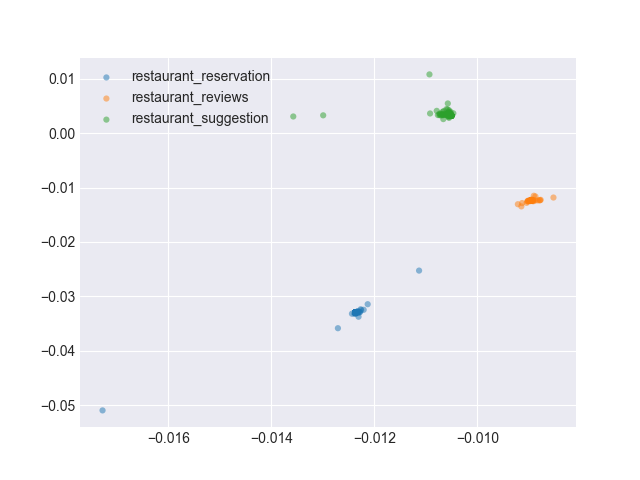}
\label{figc2}
\end{minipage}
\vspace{-0.8cm}
\caption{Visualization of in-domain representation distribution.}
\vspace{-0.4cm}
\label{vis}
\end{figure*}

\begin{table}[t]
\centering
\resizebox{0.4\textwidth}{!}{%
\begin{tabular}{|l|l|rr|rr|}
\hline
\multicolumn{2}{|l|}{settings} & \multicolumn{2}{l|}{IND} & \multicolumn{2}{l|}{OOD} \\ \hline
dimension & batch size & \multicolumn{1}{l}{ACC} & \multicolumn{1}{l|}{F1} & \multicolumn{1}{l}{Recall} & \multicolumn{1}{l|}{F1} \\ \hline
128 & 50 & 87.01 & 88.28 & 66.80 & 67.68 \\ 
128 & 100 & 87.52 & 88.60 & 67.08 & 68.12 \\ 
128 & 200 & 88.10 & 89.05 & 67.56 & 68.63 \\ \hline
256 & 50 & 87.17 & 88.40 & 66.96 & 67.92 \\ 
256 & 100 & 87.85 & 88.96 & 67.32 & 68.37 \\ 
256 & 200 & 88.37 & 89.24 & 67.76 & 68.94 \\ \hline
512 & 50 & 87.35 & 88.78 & 67.32 & 68.47 \\ 
512 & 100 & 88.14 & 89.22 & 67.64 & 68.69 \\ 
512 & 200 & \multicolumn{1}{l}{88.54} & \multicolumn{1}{l|}{89.50} & \multicolumn{1}{r}{68.00} & \multicolumn{1}{l|}{69.27} \\ \hline
\end{tabular}%
}
\vspace{-0.2cm}
\caption{Parameter analysis of batch size and representation dimension.}
\vspace{-0.6cm}
\label{app_tab2}
\end{table}

\label{few-shot}
\noindent\textbf{Effect of IND Training Data Size.} Tab \ref{tab3} shows the effect of IND training data size. We randomly choose training data with a certain proportion from CLINC-Full IND data and use the original test set for evaluation. We use the LSTM+GDA setting. Results show SCL+CE consistently outperforms CE. Besides, with the decrease of training data size, the relative improvements gradually increase. It proves SCL has strong robustness for improving OOD detection, especially in the few-shot scenario.

\label{noise}

\noindent\textbf{Analysis of Adversarial Perturbation Norm.} Fig \ref{fig3} shows the effect of adversarial perturbation norm $\epsilon$ on OOD detection performance. We conduct the experiments on CLINC-Full dataset, using LSTM and GDA. The X-axis denotes the value of $\epsilon$. The CE+GDA dashed line means no SCL pre-training and $\epsilon=0.0$ in the SCL+CE+GDA solid line means no adversarial augmentation. In general, both SCL and adversarial augmentation contribute to the improvements and $\epsilon \in (1.0, 2.0)$ achieves better performances. Compared with the baseline without SCL, the SCL+CE method with a smaller adversarial perturbation can still obtain better results but lower than the results with an optimal range of perturbation, while large norms tend to damage the effect of SCL. Our method still performs well with a broad range of adversarial perturbation and is insensitive to hyperparameters.

\noindent\textbf{Parameter Analysis.} As our proposed SCL is a method involving contrastive learning, we analyze batch sizes and representation dimensions to further verify the effectiveness, whose results are presented in Table \ref{app_tab2}. We conduct experiments in CLINC-Full dataset, using LSTM and SCL+CE objective for training and GDA for detection. With the increase of batch size and representation dimension, both in-domain and OOD metrics are slightly improved. However, compared with the method proposed in this paper, the improvement is relatively limited. In general, our proposed method is not sensitive to hyperparameters and can show the expected effect under a wide range of reasonable settings.

\noindent\textbf{Feature Visualization.} As shown in Fig \ref{vis}, we extract several groups of similar classes for PCA visualization analysis. The three pictures in the upper part represent training using only CE, while the three pictures in the lower part use SCL+CE for training. In the same column, we sample the same classes for observation. It is worth noting that the scale of the image has been adjusted adaptively in order to display all the data. The actual distance can be sensed by observing the marking of the coordinate axis. After SCL is added, the distance between similar classes is significantly expanded, and the data in the same classes are more closely clustered.

\section{Conclusion}
In this paper, we focus on the unsupervised OOD detection where no labeled OOD data exist. To learn discriminative semantic intent representations via in-domain data, we propose a novel supervised contrastive learning loss (SCL). SCL aims to minimize intra-class variance by pulling together in-domain intents belonging to the same class and maximize inter-class variance by pushing apart samples from different classes. Experiments and analysis confirm the effectiveness of SCL for OOD detection. We hope to provide new guidance for future OOD detection work.

\section*{Acknowledgements}
This work was partially supported by National Key R\&D Program of China No. 2019YFF0303300 and Subject II  No. 2019YFF0303302, DOCOMO Beijing Communications Laboratories Co., Ltd, MoE-CMCC "Artifical Intelligence" Project No. MCM20190701.

\section*{Broader Impact}
Task-oriented dialog systems have demonstrated remarkable performance across a wide range of applications, with the promise of a significant positive impact on human production mode and lifeway. However, in scenarios where information is complex and rapidly changing, models usually face input that is meaningfully different from typical examples encountered during training. Current models are prone to make unfounded predictions on these inputs, which may affect human judgment and thus impair the safety of models in practical applications. In domains with the greatest potential for societal impacts, such as navigation or medical diagnosis, models should be able to detect potentially agnostic OOD and be robust to high-entropy inputs to avoid catastrophic errors. This work proposes a novel unsupervised OOD detection method that using supervised contrastive learning to learn discriminative semantic intent representations. The effectiveness and robustness of the model are significantly improved by adding a supervised contrastive learning pre-training stage, which takes a step towards the ultimate goal of enabling the safe real-world deployment of task-oriented dialog systems in safety-critical domains. The experimental results have been reported on standard benchmark datasets for considerations of reproducible research.

\bibliographystyle{acl_natbib}
\bibliography{anthology,acl2021}

\begin{thebibliography}{33}
\expandafter\ifx\csname natexlab\endcsname\relax\def\natexlab#1{#1}\fi

\bibitem[{Akasaki and Kaji(2017)}]{Akasaki2017ChatDI}
Satoshi Akasaki and Nobuhiro Kaji. 2017.
\newblock Chat detection in an intelligent assistant: Combining task-oriented
  and non-task-oriented spoken dialogue systems.
\newblock \emph{ArXiv}, abs/1705.00746.

\bibitem[{Bendale and Boult(2016)}]{Bendale2016TowardsOS}
Abhijit Bendale and Terrance~E. Boult. 2016.
\newblock Towards open set deep networks.
\newblock \emph{2016 IEEE Conference on Computer Vision and Pattern Recognition
  (CVPR)}, pages 1563--1572.

\bibitem[{Chen et~al.(2020)Chen, Kornblith, Norouzi, and Hinton}]{Chen2020ASF}
Ting Chen, Simon Kornblith, Mohammad Norouzi, and Geoffrey~E. Hinton. 2020.
\newblock A simple framework for contrastive learning of visual
  representations.
\newblock \emph{ArXiv}, abs/2002.05709.

\bibitem[{Devlin et~al.(2019)Devlin, Chang, Lee, and
  Toutanova}]{Devlin2019BERTPO}
J.~Devlin, Ming-Wei Chang, Kenton Lee, and Kristina Toutanova. 2019.
\newblock Bert: Pre-training of deep bidirectional transformers for language
  understanding.
\newblock In \emph{NAACL-HLT}.

\bibitem[{Fei and Liu(2016)}]{Fei2016BreakingTC}
Geli Fei and Bing Liu. 2016.
\newblock Breaking the closed world assumption in text classification.
\newblock In \emph{HLT-NAACL}.

\bibitem[{Gnewuch et~al.(2017)Gnewuch, Morana, and
  Maedche}]{Gnewuch2017TowardsDC}
Ulrich Gnewuch, S.~Morana, and A.~Maedche. 2017.
\newblock Towards designing cooperative and social conversational agents for
  customer service.
\newblock In \emph{ICIS}.

\bibitem[{Goodfellow et~al.(2015)Goodfellow, Shlens, and
  Szegedy}]{goodfellow2014explaining}
Ian~J. Goodfellow, Jonathon Shlens, and Christian Szegedy. 2015.
\newblock Explaining and harnessing adversarial examples.
\newblock \emph{CoRR}, abs/1412.6572.

\bibitem[{Gunel et~al.(2020)Gunel, Du, Conneau, and
  Stoyanov}]{Gunel2020SupervisedCL}
Beliz Gunel, Jingfei Du, Alexis Conneau, and Ves Stoyanov. 2020.
\newblock Supervised contrastive learning for pre-trained language model
  fine-tuning.
\newblock \emph{ArXiv}, abs/2011.01403.

\bibitem[{He et~al.(2020)He, Fan, Wu, Xie, and Girshick}]{He2020MomentumCF}
Kaiming He, Haoqi Fan, Yuxin Wu, Saining Xie, and Ross~B. Girshick. 2020.
\newblock Momentum contrast for unsupervised visual representation learning.
\newblock \emph{2020 IEEE/CVF Conference on Computer Vision and Pattern
  Recognition (CVPR)}, pages 9726--9735.

\bibitem[{Hendrycks and Gimpel(2017)}]{Hendrycks2017ABF}
Dan Hendrycks and Kevin Gimpel. 2017.
\newblock A baseline for detecting misclassified and out-of-distribution
  examples in neural networks.
\newblock \emph{ArXiv}, abs/1610.02136.

\bibitem[{Hochreiter and Schmidhuber(1997)}]{Hochreiter1997LongSM}
S.~Hochreiter and J.~Schmidhuber. 1997.
\newblock Long short-term memory.
\newblock \emph{Neural Computation}, 9:1735--1780.

\bibitem[{Jia and Liang(2017)}]{jia2017adversarial}
Robin Jia and Percy Liang. 2017.
\newblock Adversarial examples for evaluating reading comprehension systems.
\newblock \emph{arXiv preprint arXiv:1707.07328}.

\bibitem[{Khosla et~al.(2020)Khosla, Teterwak, Wang, Sarna, Tian, Isola,
  Maschinot, Liu, and Krishnan}]{Khosla2020SupervisedCL}
Prannay Khosla, Piotr Teterwak, Chen Wang, Aaron Sarna, Yonglong Tian, Phillip
  Isola, A.~Maschinot, Ce~Liu, and Dilip Krishnan. 2020.
\newblock Supervised contrastive learning.
\newblock \emph{ArXiv}, abs/2004.11362.

\bibitem[{Kim and Kim(2018)}]{Kim2018JointLO}
Joo-Kyung Kim and Young-Bum Kim. 2018.
\newblock Joint learning of domain classification and out-of-domain detection
  with dynamic class weighting for satisficing false acceptance rates.
\newblock \emph{ArXiv}, abs/1807.00072.

\bibitem[{Kingma and Ba(2014)}]{kingma2014adam}
Diederik~P Kingma and Jimmy Ba. 2014.
\newblock Adam: A method for stochastic optimization.
\newblock \emph{arXiv preprint arXiv:1412.6980}.

\bibitem[{Kurakin et~al.(2016)Kurakin, Goodfellow, and
  Bengio}]{kurakin2016adversarial}
Alexey Kurakin, Ian Goodfellow, and Samy Bengio. 2016.
\newblock Adversarial examples in the physical world.
\newblock \emph{arXiv preprint arXiv:1607.02533}.

\bibitem[{Larson et~al.(2019)Larson, Mahendran, Peper, Clarke, Lee, Hill,
  Kummerfeld, Leach, Laurenzano, Tang, and Mars}]{larson-etal-2019-evaluation}
Stefan Larson, Anish Mahendran, Joseph~J. Peper, Christopher Clarke, Andrew
  Lee, Parker Hill, Jonathan~K. Kummerfeld, Kevin Leach, Michael~A. Laurenzano,
  Lingjia Tang, and Jason Mars. 2019.
\newblock \href {https://doi.org/10.18653/v1/D19-1131} {An evaluation dataset
  for intent classification and out-of-scope prediction}.
\newblock In \emph{Proceedings of the 2019 Conference on Empirical Methods in
  Natural Language Processing and the 9th International Joint Conference on
  Natural Language Processing (EMNLP-IJCNLP)}, pages 1311--1316, Hong Kong,
  China. Association for Computational Linguistics.

\bibitem[{Lee et~al.(2018)Lee, Lee, Lee, and Shin}]{Lee2018ASU}
Kimin Lee, Kibok Lee, Honglak Lee, and Jinwoo Shin. 2018.
\newblock A simple unified framework for detecting out-of-distribution samples
  and adversarial attacks.
\newblock \emph{ArXiv}, abs/1807.03888.

\bibitem[{Lin and Xu(2019)}]{Lin2019DeepUI}
Ting-En Lin and Hua Xu. 2019.
\newblock Deep unknown intent detection with margin loss.
\newblock In \emph{ACL}.

\bibitem[{Pennington et~al.(2014)Pennington, Socher, and
  Manning}]{pennington2014glove}
Jeffrey Pennington, Richard Socher, and Christopher~D Manning. 2014.
\newblock Glove: Global vectors for word representation.
\newblock In \emph{Proceedings of the 2014 conference on empirical methods in
  natural language processing (EMNLP)}, pages 1532--1543.

\bibitem[{Ren et~al.(2019)Ren, Liu, Fertig, Snoek, Poplin, DePristo, Dillon,
  and Lakshminarayanan}]{Ren2019LikelihoodRF}
Jie Ren, Peter~J. Liu, Emily Fertig, Jasper Snoek, Ryan Poplin, Mark~A.
  DePristo, Joshua~V. Dillon, and Balaji Lakshminarayanan. 2019.
\newblock Likelihood ratios for out-of-distribution detection.
\newblock \emph{ArXiv}, abs/1906.02845.

\bibitem[{Rozsa et~al.(2016)Rozsa, Rudd, and Boult}]{rozsa2016adversarial}
Andras Rozsa, Ethan~M Rudd, and Terrance~E Boult. 2016.
\newblock Adversarial diversity and hard positive generation.
\newblock In \emph{Proceedings of the IEEE Conference on Computer Vision and
  Pattern Recognition Workshops}, pages 25--32.

\bibitem[{Scheirer et~al.(2013)Scheirer, Rocha, Sapkota, and
  Boult}]{Scheirer2013TowardOS}
Walter~J. Scheirer, Anderson Rocha, Archana Sapkota, and Terrance~E. Boult.
  2013.
\newblock Toward open set recognition.
\newblock \emph{IEEE Transactions on Pattern Analysis and Machine
  Intelligence}, 35:1757--1772.

\bibitem[{Shu et~al.(2017)Shu, Xu, and Liu}]{Shu2017DOCDO}
Lei Shu, Hu~Xu, and Bing Liu. 2017.
\newblock Doc: Deep open classification of text documents.
\newblock In \emph{EMNLP}.

\bibitem[{Shum et~al.(2018)Shum, He, and Li}]{Shum2018FromET}
H.~Shum, X.~He, and Di~Li. 2018.
\newblock From eliza to xiaoice: challenges and opportunities with social
  chatbots.
\newblock \emph{Frontiers of Information Technology \& Electronic Engineering},
  19:10--26.

\bibitem[{Tulshan and Dhage(2018)}]{Tulshan2018SurveyOV}
Amrita~S. Tulshan and S.~Dhage. 2018.
\newblock Survey on virtual assistant: Google assistant, siri, cortana, alexa.

\bibitem[{Wang et~al.(2018)Wang, Cheng, Liu, and Liu}]{Wang2018AdditiveMS}
Feng Wang, Jian Cheng, Weiyang Liu, and H.~Liu. 2018.
\newblock Additive margin softmax for face verification.
\newblock \emph{IEEE Signal Processing Letters}, 25:926--930.

\bibitem[{Xu et~al.(2020)Xu, He, Yan, Liu, Liu, and Xu}]{xu-etal-2020-deep}
Hong Xu, Keqing He, Yuanmeng Yan, Sihong Liu, Zijun Liu, and Weiran Xu. 2020.
\newblock \href {https://www.aclweb.org/anthology/2020.coling-main.125} {A deep
  generative distance-based classifier for out-of-domain detection with
  mahalanobis space}.
\newblock In \emph{Proceedings of the 28th International Conference on
  Computational Linguistics}, pages 1452--1460, Barcelona, Spain (Online).
  International Committee on Computational Linguistics.

\bibitem[{Yan et~al.(2020)Yan, He, Xu, hong Liu, Meng, Hu, and
  Xu}]{Yan2020AdversarialSD}
Yuanmeng Yan, Keqing He, H.~Xu, Si~hong Liu, Fanyu Meng, Min Hu, and Weiran Xu.
  2020.
\newblock Adversarial semantic decoupling for recognizing open-vocabulary
  slots.
\newblock In \emph{EMNLP}.

\bibitem[{Zeng et~al.(2021{\natexlab{a}})Zeng, He, Yan, Xu, and
  Xu}]{Zeng2021AdversarialSL}
Zhiyuan Zeng, Keqing He, Yuanmeng Yan, Hong Xu, and Weiran Xu.
  2021{\natexlab{a}}.
\newblock Adversarial self-supervised learning for out-of-domain detection.
\newblock In \emph{NAACL}.

\bibitem[{Zeng et~al.(2021{\natexlab{b}})Zeng, Xu, He, Yan, Liu, Liu, and
  Xu}]{9413908}
Zhiyuan Zeng, Hong Xu, Keqing He, Yuanmeng Yan, Sihong Liu, Zijun Liu, and
  Weiran Xu. 2021{\natexlab{b}}.
\newblock \href {https://doi.org/10.1109/ICASSP39728.2021.9413908} {Adversarial
  generative distance-based classifier for robust out-of-domain detection}.
\newblock In \emph{ICASSP 2021 - 2021 IEEE International Conference on
  Acoustics, Speech and Signal Processing (ICASSP)}, pages 7658--7662.

\bibitem[{Zhang et~al.(2019)Zhang, Zhou, Miao, and Li}]{zhang2019generating}
Huangzhao Zhang, Hao Zhou, Ning Miao, and Lei Li. 2019.
\newblock Generating fluent adversarial examples for natural languages.
\newblock In \emph{Proceedings of the 57th Annual Meeting of the Association
  for Computational Linguistics}, pages 5564--5569.

\bibitem[{Zheng et~al.(2020)Zheng, Chen, and Huang}]{Zheng2020OutofDomainDF}
Yinhe Zheng, Guanyi Chen, and Minlie Huang. 2020.
\newblock Out-of-domain detection for natural language understanding in dialog
  systems.
\newblock \emph{IEEE/ACM Transactions on Audio, Speech, and Language
  Processing}, 28:1198--1209.

\end{thebibliography}

\clearpage

\appendix

\section{Dataset Details}
\label{datasetdetails}

Table \ref{dataset} shows the details of two benchmark OOD dataset\footnote{https://github.com/clinc/oos-eval} CLINC-Full and CLINC-Small \cite{larson-etal-2019-evaluation}. They both contain 150 in-domain intents across 10 domains. It is worth noting that our paper does not use labeled OOD data from the training set in the training stage.

\begin{table}[h]
\centering
\resizebox{0.48\textwidth}{!}{%
\begin{tabular}{l|cc}
\hline
CLINC               & Full       & Small \\ \hline
Avg utterance length & 9          & 9           \\
Intents              & 150        & 150         \\
Training set size    & 15100      & 7600        \\
Training samples per class  & 100      & 50 \\
Training OOD samples amount & 100      & 100 \\
Development set size & 3100       & 3100        \\
Development samples per class & 20       & 20        \\
Development OOD samples amount & 100       & 100 \\
Testing Set Size     & 5500       & 5500        \\
Testing samples per class & 30       & 30        \\
Development OOD samples amount & 1000       & 1000 \\ \hline
\end{tabular}
}
\caption{Statistics of the CLINC datasets.}
\label{dataset}
\end{table}

\begin{table*}[t]
\centering
\resizebox{0.9\textwidth}{!}{%
\begin{tabular}{l|l|rr|rr|rr|rr}
\hline
\multicolumn{2}{l|}{\multirow{3}{*}{Models}} & \multicolumn{4}{l|}{CLINC-Full} & \multicolumn{4}{l}{CLINC-Small} \\ \cline{3-10} 
\multicolumn{2}{l|}{} & \multicolumn{2}{l|}{IND} & \multicolumn{2}{l|}{OOD} & \multicolumn{2}{l|}{IND} & \multicolumn{2}{l}{OOD} \\ \cline{3-10} 
\multicolumn{2}{l|}{} & \multicolumn{1}{l}{ACC} & \multicolumn{1}{l|}{F1} & \multicolumn{1}{l}{Recall} & \multicolumn{1}{l|}{F1} & \multicolumn{1}{l}{ACC} & \multicolumn{1}{l|}{F1} & \multicolumn{1}{l}{Recall} & \multicolumn{1}{l}{F1} \\ \hline
\multirow{4}{*}{LOF} & CE & 85.46 & 85.80 & 57.40 & 58.78 & 82.45 & 82.73 & 52.88 & 53.90 \\ 
 & LMCL & 85.87 & 86.08 & 58.32 & 59.28 & 82.83 & 82.98 & 53.96 & 54.63 \\ \cline{2-10} 
 & SCL+CE(ours) & 86.52 & 86.80 & 60.72 & 61.80 & 83.13 & 83.39 & 56.88 & 57.48 \\  
 & SCL+LMCL(ours) & \textbf{86.94} & \textbf{87.15} & \textbf{61.88} & \textbf{63.03} & \textbf{83.40} & \textbf{83.57} & \textbf{57.92} & \textbf{58.60} \\ \hline
\multirow{4}{*}{MSP} & CE & 85.76 & 86.27 & 27.12 & 34.91 & 83.81 & 84.12 & 20.40 & 22.76 \\ 
 & LMCL & 87.36 & 87.62 & 31.28 & 36.66 & 85.02 & 85.30 & 24.16 & 25.72 \\ \cline{2-10} 
 & SCL+CE(ours) & 87.44 & 87.87 & 33.68 & 39.34 & 85.54 & 85.95 & 27.24 & 27.43 \\ 
 & SCL+LMCL(ours) & \textbf{88.89} & \textbf{89.21} & \textbf{35.40} & \textbf{41.75} & \textbf{86.87} & \textbf{87.20} & \textbf{29.28} & \textbf{31.02}  \\ \hline
\end{tabular}%
}
\caption{Supplementary experimental results of LOF and MSP.}
\label{app_tab1}
\end{table*}


\begin{table}[]
\centering
\resizebox{0.40\textwidth}{!}{%
\begin{tabular}{|l|ll|ll|}
\hline
\multirow{2}{*}{models} & \multicolumn{2}{l|}{IND} & \multicolumn{2}{l|}{OOD} \\ \cline{2-5} 
 & ACC & F1 & Recall & F1 \\ \hline
CE & 86.34 & 87.73 & 63.72 & 65.23 \\ \hline
CE+SCL & 82.29 & 83.59 & 61.96 & 63.40 \\ \hline
multitask & 86.69 & 88.02 & 65.76 & 67.25 \\ \hline
SCL+CE & \textbf{87.01} & \textbf{88.28} & \textbf{66.80} & \textbf{67.68} \\ \hline
\end{tabular}%
}
\caption{Results of combining two training stages in different ways}
\label{app_tab3}
\end{table}

\section{Baseline Details}
We compare many types of unsupervised OOD detection models. Therefore, the model proposed in this paper can be divided into the training stage and detection stage. For each model LSTM or BERT, we use different detection methods to verify its performance. The innovation of this paper focuses mainly on the training stage. Due to the limitation of space, we do not detailed introduce the detection methods in the main body. We will supplement the relevant contents as follows: 

\noindent\textbf{MSP} (Maximum Softmax Probability)\cite{Hendrycks2017ABF} applies a threshold on the maximum softmax probability where the threshold is set to 0.5 according to the dev set. 

\noindent\textbf{LOF} (Local Outlier Factor)\cite{Lin2019DeepUI} uses the local outlier factor to detect unknown intents. The motivation is that if an example’s local density is significantly lower than its k-nearest neighbor’s, it is more likely to be considered as the unknown intents.

\noindent\textbf{GDA} (Gaussian Discriminant Analysis)\cite{xu-etal-2020-deep} is a generative distance-based classifier for out-of-domain detection with Euclidean space. They estimate the class-conditional distribution on feature spaces of DNNs via Gaussian discriminant analysis (GDA) to avoid over-confidence problems and use Mahalanobis distance to measure the confidence score of whether a test sample belongs to OOD. GDA is the state-of-the-art detection methods till now, so we adopt GDA as our main detection algorithm. We also report MSP and LOF results in Section \ref{main}.

\section{Implementation Details}
We use the public pre-trained 300 dimensions GloVe embeddings \cite{pennington2014glove}\footnote{https://github.com/stanfordnlp/GloVe} or bert-base-uncased \cite{Devlin2019BERTPO}\footnote{https://github.com/google-research/bert} model to embed tokens. We use a single-layer BiLSTM as a feature extractor and set the dimension of hidden states to 128. The dropout value is fixed at 0.5. We use Adam optimizer \cite{kingma2014adam} to train our model. We set a learning rate to 1E-03 for GloVe+LSTM and 1E-04 for Bert. In the training stage, 100 epochs of supervised contrastive training are first conducted, then 10 epochs of finetune training are conducted with CE or LMCL. Both phases are training only on in-domain labeled data. The training stage has an early stop setting with patience equal to 5. We use the best F1 scores on the validation set to calculate the GDA threshold adaptively. Each result of the experiments is tested 5 times under the same setting and gets the average value. The norms of adversarial perturbation are obtained by the heuristic method, in which MSP and LOF are 1.0 and GDA is 1.5. The training stage of our model lasts about 10 minutes using GloVe embeddings, and 18 minutes using Bert-base-uncased, both on a single Tesla T4 GPU(16 GB of memory). The average value of the trainable model parameters is 3.05M. 

\section{Supplementary Experimental Results}
\label{main}
\noindent\textbf{Various Detection Methods} In this paper, the experiments and analysis are mainly conducted around the training stage. Different detection models are used to verify the generalization of our proposed method. Due to the limitation of space, we use GDA for most of the presentation in the main body. The main experiments of LOF and MSP using LSTM feature extractor are shown in Table \ref{app_tab1}. It is worth noting that using different detection methods can obtain the same analysis results as the main experimental in the main body.



\noindent\textbf{Combining two training stages in different ways} We display results of different combining ways of two training stages on CLINC-Full dataset using LSTM and GDA detection method in Table \ref{app_tab3}. CE is the baseline that only uses the cross-entropy loss function to train the feature extractor. SCL+CE follows the paradigm of pre-training first and then finetuning, which achieves the best performance. Besides, we try two different combinations to explore the relationship between the two training stages. CE+SCL means that we first conduct training to minimize cross-entropy loss, and then conduct supervised contrastive learning. The results show that the subsequent SCL leads to a decline in metrics, especially on in-domain. This is because SCL, while optimizing the representation distribution, compromises the mapping relationship with labels. Multitask means to optimize two losses simultaneously. This setting leads to mutual interference between two tasks, which affects the convergence effect and damages the performance and stability of the model. In general, SCL should be used as a pre-training method and CE as a finetuning method. The best results can be achieved by first using SCL to learn discriminative representation and then finetuning the model by CE.

\end{document}